\documentclass[conference]{IEEEtran}

\usepackage{cite}
\usepackage{amsmath,amssymb,amsfonts}
\usepackage{algorithmic}
\usepackage{graphicx}
\usepackage{textcomp}
\usepackage{xcolor}
\usepackage{times}
\usepackage{epsfig}
\usepackage{url}
\usepackage{booktabs}
\usepackage{blindtext}
\usepackage{caption}
\usepackage[bookmarks=false]{hyperref}

\newcommand{\algoname}{\emph{Shooting Labels}}

\begin{document}

\title{Shooting Labels: 3D Semantic Labeling\\by Virtual Reality}

\author{\IEEEauthorblockN{Pierluigi Zama Ramirez\textsuperscript{1},  Claudio Paternesi\textsuperscript{1}, Luca De Luigi\textsuperscript{1}, Luigi Lella\textsuperscript{1}, Daniele De Gregorio\textsuperscript{2}, Luigi Di Stefano\textsuperscript{1}}
\IEEEauthorblockA{
\textsuperscript{1}Department of Computer Science and Engineering (DISI) - University of Bologna, Italy\\
\textsuperscript{2}Eyecan.ai\\
{\tt\small \{pierluigi.zama, luca.deluigi4,  luigi.distefano\}@unibo.it}\\
{\tt\small \{claudio.paternesi, luigi.lella\}@studio.unibo.it} \\
{\tt\small daniele.degregorio@eyecan.ai} 
}}


\twocolumn[{%
\renewcommand\twocolumn[1][]{#1}%
\maketitle
\begin{center}
    \centering
    \includegraphics[width=\textwidth]{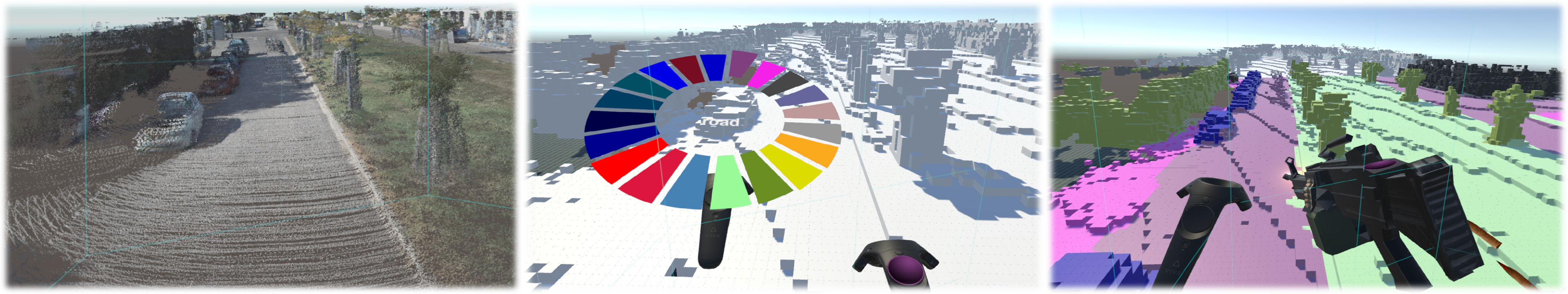} 
    \captionof{figure}{\textbf{V}irtual \textbf{R}eality view of a 3D reconstruction from KITTI sequence. Left image: RGB visualization of the virtual world. Middle image: label palette and empty voxelization. Right image: partially labeled environment.}
    \label{fig:teaser}
\end{center}
}]

\begin{abstract}
Availability of a few, large-size, annotated datasets, like ImageNet, Pascal VOC and COCO, has lead deep learning to revolutionize computer vision research by achieving astonishing results in several  vision tasks. We argue that new tools to facilitate generation of annotated datasets may help spreading data-driven AI throughout applications and domains.
In this work we propose Shooting Labels, the first 3D labeling tool for dense 3D semantic segmentation which exploits Virtual Reality to render the labeling task as easy and fun as playing a video-game. Our tool allows for semantically labeling large scale environments very expeditiously, whatever the nature of the 3D data at hand (e.g. point clouds, mesh).
Furthermore, Shooting Labels efficiently integrates multi-users annotations to improve the labeling accuracy automatically and compute  a label uncertainty map. Besides, within our framework the 3D annotations can be projected into 2D images, thereby  speeding up also a notoriously slow and expensive task such as pixel-wise semantic labeling. 
We demonstrate the accuracy and efficiency of our tool in two different scenarios: an indoor workspace provided by Matterport3D and a large-scale outdoor environment reconstructed from 1000+ KITTI images.
\end{abstract}

\begin{figure*}[ht]
	\includegraphics[width=\textwidth]{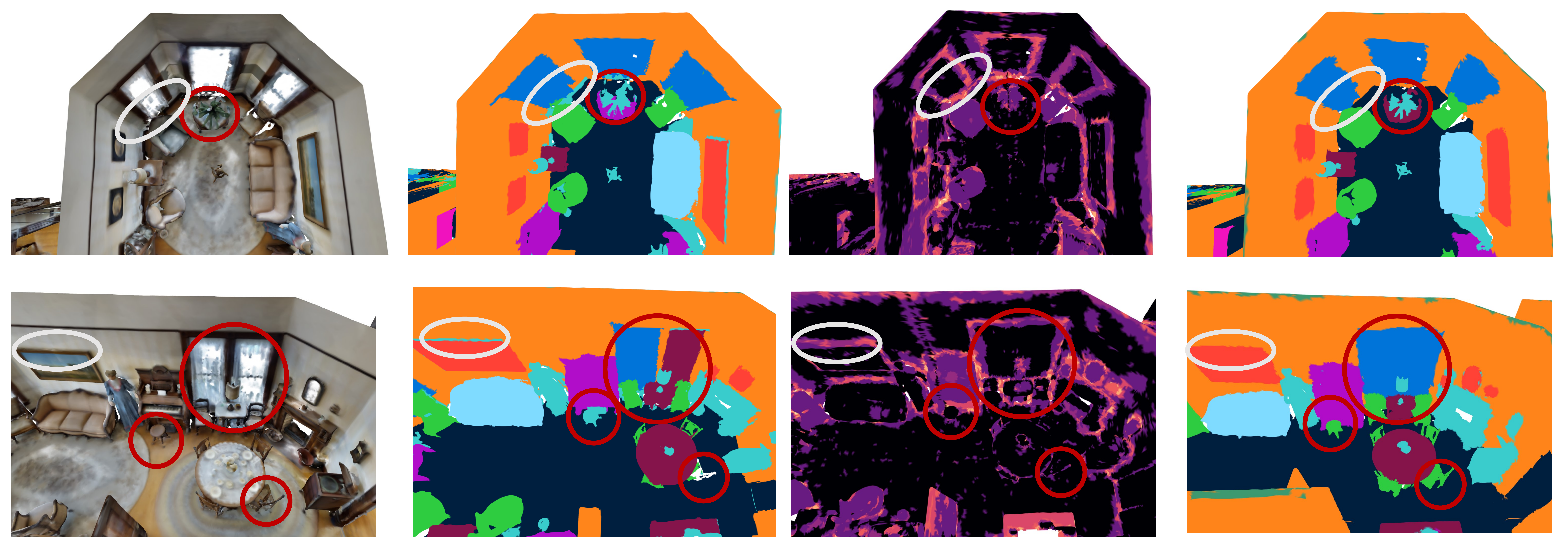}
	\caption{Matterport 3D dataset. From left to right: RGB mesh, ground-truth provided by Matterport, Uncertainty map by multi-user integration, best results obtained with our tool. Circled in red we highlight the errors in the Matterport ground-truth (table as furniture, windows as table) avoided by labeling with our tool. Circled in white we show high uncertainty labels (e.g. 3D boundaries).}
	\label{fig:gt_errors}
\end{figure*}

\section{Introduction}
Two major leitmotifs in nowadays computer vision are: \emph{Convolutional Neural Networks have surpassed human performance in classification tasks} \cite{he2015delving} and \emph{The success of the modern Deep Neural Networks (DNNs) is ascribable to the availability of large datasets} \cite{cordts2016cityscapes}. 

As for the latter, one might just consider the dramatic advances brought in by large annotated datasets like ImageNet \cite{russakovsky2015imagenet} and Pascal VOC \cite{everingham2015pascal} in the fields of image classification and object detection, as well as by KITTI \cite{geiger2013vision} and Cityscapes \cite{cordts2016cityscapes} in the realm of dense scene understanding. 
Indeed, the key issue in modern computer vision deals more and more with how to speed-up and facilitate acquisition of large annotated datasets. 
Innovative companies, like Scale.ai\footnote{https://scale.ai/}, 
Superannotate.ai\footnote{https://superannotate.ai/}
and many others, are gaining increasing popularity thanks to their  advanced image labeling tools.
This suggests data annotation to play a pivotal role alongside with the development of data-driven AI techniques.

The annotation process is notoriously tedious and expensive.
Moreover, the more complex the visual perception task, the slower and more costly becomes the required  annotation work. 
For instance, labeling a single image for 2D semantic segmentation, one of the most complex annotation tasks together with instance segmentation, can take several hours per image.
Thus, as proposed in  \cite{hua2016scenenn,dai2017scannet,mccormac2017scenenet,Matterport3D}, directly annotating a 3D reconstruction of the scene and then projecting the 3D labels into 2D images can facilitate the data generation process significantly.
On the other hand, the annotation process for 3D Dense Semantic Segmentation requires  expertise in 3D modeling and several consecutive hours of manual labeling, which makes it an error-prone process, also due to inherent difficulties in handling 3D data (occlusions, 3D boundaries, ambiguities  introduced by the 3D reconstruction process). 
Thus, existing 3D labeling tools \cite{chang2017matterport3d, dai2017scannet, mccormac2017scenenet} employ several annotators to iteratively refine the labeled  data and attain  the so-called \emph{ground truth}, which, nonetheless, turns out far from perfect.
For instance, in \autoref{fig:gt_errors} we can see exemplar annotation errors (e.g. a window labeled as a table).
The majority of annotated datasets, both 2D and 3D, are obtained through similar procedures and thus may show some mistakes, leading to biased rankings when employed to benchmark the accuracy of algorithms.

Based on these considerations, in this work we propose \algoname{}, a novel tool based on Virtual Reality (VR) to ease the dense 3D semantic labeling, so as to gather 3D and 2D data endowed with semantic annotations.  To the best of our knowledge, ours is the first system which allows for handling efficiently large-scale 3D semantic labeling processes, such as labeling whole city blocks. Moreover, by exploiting Virtual Reality to make the task of labeling as easy and fun as playing a video-game, our approach remarkably reduces the expertise necessary to work with  3D semantic labeling tools. 
The immersive experience provided by VR technologies allows the user to physically move around within the scenario she/he is willing  to label and interact with objects in a natural and engaging way. The user is transported into a large virtual environment represented as 3D meshes, where surfaces can be ``colored'' semantically in a highly captivating way (see. \autoref{fig:teaser} for some in-game visualizations).

The full fledged gamification of our tool empowers a larger community to undertake this type of activity and enables the possibility to obtain much more annotated data. For this reason, our tool features a  multi-player post-processing procedure wherein we integrate results of several annotators to both improve accuracy and compute a labeling uncertainty map which provides information about  the reliability of the produced  ground truth.

The novel contributions of this paper can thus be summarized as follows:

\begin{itemize}
	\item To the best of our knowledge, we  propose the first 3D dense labeling tool based on Virtual Reality.
	\item Our immersive game-style interface enables users without specific expertise to label 3D data expeditiously.
	\item Our tool can work with the most common 3D data structures and with large scale scenarios.
	\item We propose an approach to integrate multi-player results in order to extract information about the reliability of the generated ground truth. Thus, unlike existing tools, our provide both semantic labels and their associated uncertainty.
	\item We label a whole sequence of KITTI and an entire house of Matterport3D and release annotations dealing with both 3D data and 2D images.
\end{itemize}

Our open-source framework is based on Unity\footnote{https://unity.com/}, Blender\footnote{https://www.blender.org/} and open3D \cite{Zhou2018}. Project page:  \color{purple} \url{https://github.com/pierlui92/Shooting-Labels}\color{black}.

\begin{figure*}
	\centering
	\includegraphics[width=0.94\textwidth]{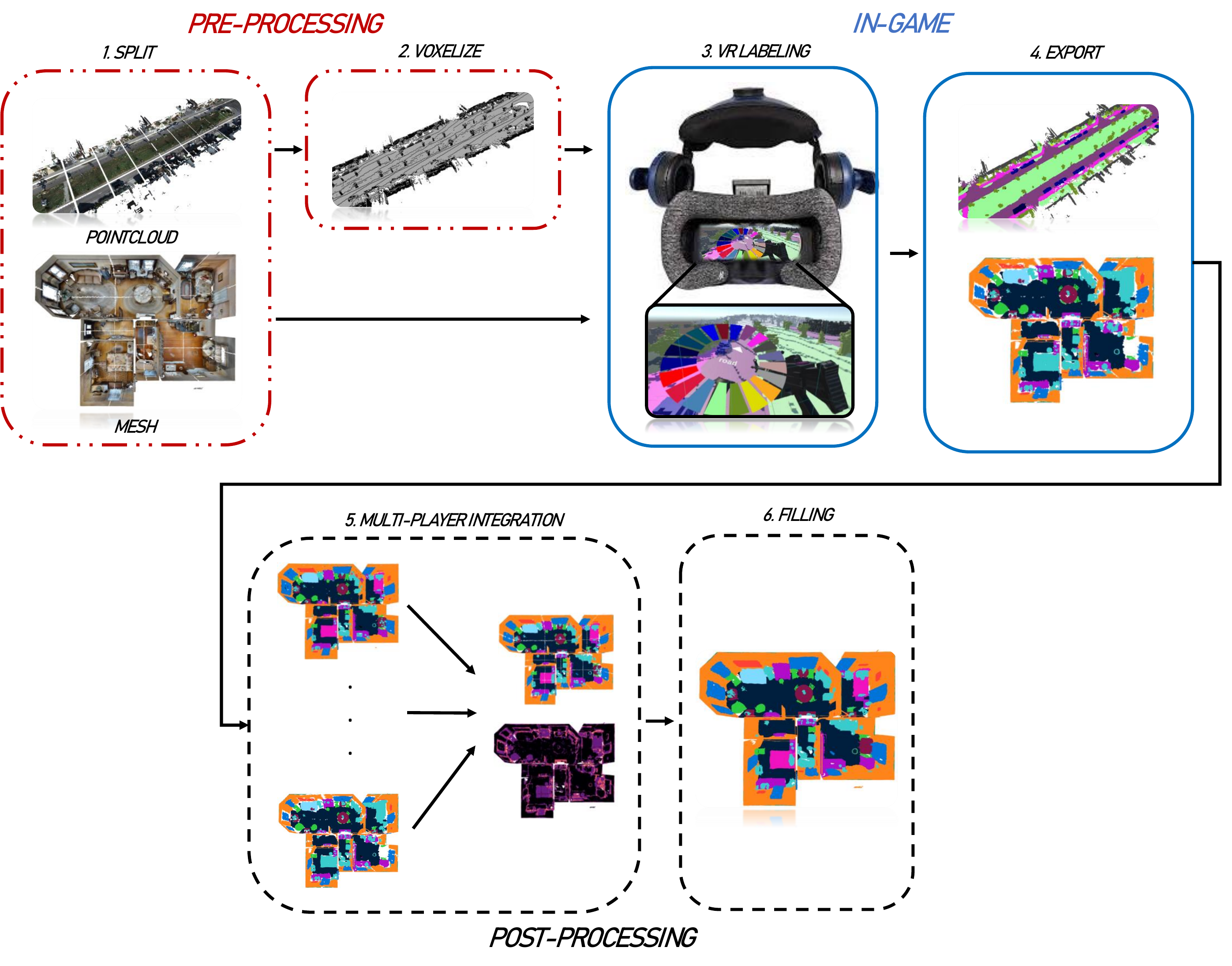}\\
	\caption{The six steps of the \algoname{} pipeline. 1- Splitting 3D data (a mesh or point cloud) into chunks to optimize visualization in the VR environment. 2- In case of point clouds, voxelization enables real-time rendering and reduces memory footprint 3- Semantic labeling by Virtual Reality. 4- Exporting annotated data into the original format 5- Integrating multi-player results to improve labeling and assess the reliability of the labels (optional). 6- Filling of unlabeled elements (optional).}
	\label{fig:pipeline}
\end{figure*}

\section{Related Works}
We review the literature concerning Semantic Segmentation datasets and the user experience associated with 3D dense annotation.

\textbf{Semantic Segmentation Datasets}
Several datasets featuring 2D images annotated with semantic labels are available. The most popular  are  KITTI \cite{geiger2013vision} and Cityscapes \cite{cordts2016cityscapes}, which, yet, contain a relatively small number of  images semantically annotated by hand. They were two of the first datasets proposed in this area (urban outdoor), so the focus was more on the data than on how to generate them. On the other hand, the Mapillary dataset \cite{neuhold2017mapillary} includes many more images, though the labeling was still performed manually image by image. The same is true for some indoor datasets, such as \cite{silberman2012indoor} and \cite{song2015sun}, in which, although smart graphical tools are used to produce  frame-by-frame annotations, the order of magnitude of the available images is only slightly higher. Conversely, in \cite{hua2016scenenn} an efficient pipeline for indoor environments is proposed, which allows for scanning a room, reconstruct it in 3D  and then easily label  the gathered 3D data rather than each single image. In \cite{dai2017scannet} such procedure is formally extended with a projection module which, based on known camera poses, brings the 3D labels into 2D. The label projection approach was then exploited in other datasets, such as \cite{armeni2017joint} and \cite{Matterport3D}. It can be observed that, leveraging on 3D reconstruction and camera tracking to facilitate labeling, may be thought of as shifting the cost of labeling each individual image toward  the complexity of the requirements necessary to obtain a suitable dataset (tracked camera). This benefit is even more evident in  synthetic datasets, such as \cite{handa2016understanding, mccormac2017scenenet, InteriorNet18}, where obviously both camera tracking and 3D reconstruction are no longer external elements but inherent to the rendering engine.
Recently \cite{Huang_2018_CVPR_Workshops, hackel2017isprs,behley2019dataset} have proposed large urban outdoor 2D-3D datasets where the annotation task is performed on point clouds and semantically labeled images are attained by projection.

\textbf{User Experience}
As for  ``how to generate the data'', the user experience associated with 3D Dense Annotation is rarely addressed in literature. Indeed, this procedure typically involves professional software and 3D modeling experience. Some authors have expressly addressed this by proposing elegant, efficient solutions. In \cite{valentin2015semanticpaint} the authors have proposed an interactive procedure by means of which the user can physically touch the scene within a classical 3D reconstruction pipeline, so as to ``color'' large parts of the scene by exploiting region growing techniques. In \cite{miksik2015semantic}, instead, the authors build a physical device able to reproduce the pipeline while the user navigates the environment in Augmented Reality, using a laser pointer to identify the homogeneous areas of the scene and assign them a correct label.
Differently, our proposal introduces a Virtual Reality framework to navigate within the reconstructed environments, providing the user with a series of gamification tools  to ``color'' the world in a fast and intuitive manner. To the best of our knowledge, ours is the first method that allows for labeling very-large-scale scenes in a short time by a VR approach. 

\section{VR Labeling Tool}
In this section we describe the key features of our tool. \algoname{} works with the most popular 3D representations, such as point cloud and meshes, which can be obtained by any kind of 3D reconstruction technique. Moreover, with our tool we can also load a 3D 
scene pre-labeled by any other technique (e.g. a CNN for 3D semantic segmentation) in order to refine it. As shown in \autoref{fig:pipeline}, our pipeline can be summarized into 6 main steps grouped into 3 stages, with the first stage dealing with \emph{Pre-processing} of the input 3D Data, the second with \emph{In-Game} labeling  and the third with  \emph{Post-processing} of the labeled 3D data.


\subsection{Pre-Processing of the Input 3D Data}
Meshes and point clouds obtained by 3D reconstruction techniques typically consist of millions of vertices which can hardly be rendered in real-time in a VR environment. For this reason, with both meshes and point clouds we employ a Level Of Detail strategy to mitigate the computational demand. As illustrated in the first step of \autoref{fig:pipeline}, we split meshes and point clouds in several chunks, saving each chunk at 3 different resolutions. During VR labeling session, objects close to the player are loaded at a higher resolution than those far away.
Furthermore, as point clouds cannot be managed by the Unity gaming engine,  we voxelize them (see \autoref{fig:pipeline}) in order to obtain a friendly visualization both in terms of light computation and user-experience during navigation.
To perform voxelization we set a discretization step and for each position of the dense 3D grid  build a cube mesh if that volume contains a  minimum number of points (e.g. $>$ 5).

\begin{table*}[!htbp]
	\center
	\begin{tabular}{c|ccccccccccc|cc}
		\toprule
		\rotatebox{90}{Player} &       \rotatebox{90}{Bed} & 	    \rotatebox{90}{Ceiling} & \rotatebox{90}{Chair} & 	\rotatebox{90}{Floor} &     \rotatebox{90}{Furniture} & \rotatebox{90}{Object} & 	\rotatebox{90}{Picture} &   \rotatebox{90}{Sofa} & 	    \rotatebox{90}{Table} & 	\rotatebox{90}{Wall} & 	    \rotatebox{90}{Window} &     \rotatebox{90}{Perc.Labeled} & \rotatebox{90}{mIoU}\\
		\midrule 
		1 & 74.63 &  \textbf{95.19} & 69.35 & 84.16 & 78.83 & 58.66 & 74.74 & 85.99 & 55.14 & 86.86 & 71.84 & \textbf{97.04} & 75.94\\ 
		2 & 76.37 & 90.29 & 72.45 &  \textbf{88.16} &  \textbf{80.87} & 60.76 & \textbf{77.36} & 82.39 & \textbf{62.20} & 85.65 & 69.43 & 94.56 & 76.90\\ 
		3 & 74.15 & 93.53 & 71.19 & 80.28 & 67.14 & 50.12 & 66.80 & 80.43 & 56.56 & 86.31 & 70.02 & 94.31 & 72.41\\ 
		4 &   -   & 90.07 & 70.74 & 80.12 & 58.78 & 44.89 & 64.32 & 80.60 & 38.79 & 84.72 & 68.93 & 88.44 & 62.00\\ 
		5 & 77.69 & 89.10 & 35.99 & 77.76 & 56.09 & 41.87 & 46.90 & 33.36 & 40.49 & 80.95 & 44.25 & 91.76 & 56.77\\ 
		Integr. & \textbf{81.46} & 94.80 & \textbf{77.34} & 87.02 & 79.46 & \textbf{64.27} & 75.35 & \textbf{89.68} & 57.70 & \textbf{89.83} & \textbf{74.93} & 94.47 & \textbf{79.26}\\ 
		\bottomrule
	\end{tabular}
	\small
	\caption{Comparison between single player and multi-player results on the Matterport 3D dataset \cite{Matterport3D}. Best results in bold.}
	\label{tab:singlevsmulti}
\end{table*}

\begin{table*}[!htbp]
	\center
	\begin{tabular}{c|ccccccccccc|cc}
		\toprule
		\rotatebox{90}{Player} &       \rotatebox{90}{Bed} & 	    \rotatebox{90}{Ceiling} & \rotatebox{90}{Chair} & 	\rotatebox{90}{Floor} &     \rotatebox{90}{Furniture} & \rotatebox{90}{Object} & 	\rotatebox{90}{Picture} &   \rotatebox{90}{Sofa} & 	    \rotatebox{90}{Table} & 	\rotatebox{90}{Wall} & 	    \rotatebox{90}{Window} &     \rotatebox{90}{Perc.Labeled} & \rotatebox{90}{mIoU}\\
		\midrule 
		1 & 73.77 & \textbf{95.18} & 68.09 & 82.94 & 78.04 & 56.28 & 73.79 & \textbf{85.56} & 54.29 & 86.13 & 71.58 & 100 &  75.06 \\ 
		2 & 75.90 & 90.28 & 65.55 & \textbf{84.69} & \textbf{78.31} & 55.44 & \textbf{76.00} & 67.00 & \textbf{58.60} & 84.05 & 68.19 & 100 & 73.09\\ 
		3 & 73.21 & 93.23 & 67.00 & 77.54 & 65.27 & 48.08 & 65.57 & 75.25 & 52.64 & 84.46 & 68.68 & 100 & 70.09\\ 
		4 &   -   & 88.86 & 57.47 & 73.92 & 54.73 & 39.56 & 59.50 & 68.01 & 30.52 & 80.93 & 65.67 & 100 & 56.29\\ 
		5 & 71.56 & 88.82 & 35.13 & 73.79 & 54.49 & 38.55 & 46.13 & 32.30 & 38.50 & 78.59 & 43.89 & 100 & 54.71\\ 
		Integr. & \textbf{80.33} & 94.70 & \textbf{70.27} & 83.00 & 77.90 & \textbf{57.98} & 74.51 & 82.94 & 53.86 & \textbf{87.85} & \textbf{73.85} & 100 & \textbf{76.11}\\ 
		\midrule
		Integr.0.5  & {\color{red}80.36} & 94.71 & {\color{red}70.30} & {\color{red}83.02} & {\color{red}77.94} & {\color{red}57.99} & 74.58 & 82.86 & {\color{red}53.95} & 87.87 & 73.86 & 100 & {\color{red}76.13}\\ 
		Integr.0.65 & 77.97 & {\color{red}94.85} & 70.15 & 82.13 & 77.84 & 57.24 & {\color{red}74.71} & {\color{red}85.28} & 50.65 & {\color{red}87.91} & {\color{red}74.78} & 100 & 75.77\\ 
		Integr.0.8  & 75.17 & 94.40 & 68.62 & 81.10 & 77.91 & 55.62 & 73.21 & 81.91 & 47.40 & 87.13 & 70.27 & 100 &  73.89\\ 
		\bottomrule
	\end{tabular}
	\small
	\caption{Comparison between filling single users, multi-player integration and multi-player integration based on label uncertainty on the Matterport 3D dataset \cite{Matterport3D}. Best results of filling without uncertainty in bold. Best result of filling integration using uncertainty in red.}
	\label{tab:filling}
\end{table*}

\subsection{In-Game Labeling}
3D meshes are loaded into the Virtual World and the user can explore and label the environment. The player can teleport or physically move around the scene to reach each portion of the environment. The following features have been implemented to enhance and simplify the user experience:
\begin{itemize}
	\itemsep0em 
	\item Geometric and RGB visualization
	\item Unlabeled Face Visualization
	\item Level of Detail (LOD)
	\item Labeling Granularity
	\item Export of Final Results
\end{itemize}

\textbf{Geometric and RGB Visualization}
To assign a semantic label to a mesh, the user paints on the geometric view of the object (\autoref{fig:teaser} middle picture).
However, in the  3D reconstruction objects may be difficult to disambiguate without color cues. To address this, in the case of meshes, we directly visualize the RGB version of the mesh if available.
As for point clouds, we noticed that coarse RGB voxelizations can lead the user to misunderstand the scene.
Thus, we visualize directly the RGB point cloud, building a mesh object for each point to enable visualization of this type of data also within the Unity rendering system (\autoref{fig:teaser} left image). We did not employ this kind of visualization during labeling because the interaction with this type of data can be extremely slow. However, we obtain smooth rendering performances for only the visualization. 

\textbf{Unlabeled Face Visualization}
Reaching some portion of the 3D space can be hard (e.g small hidden faces), or the user might wish to visualize the progress of its labeling. Thus, we keep track of the faces labeled by the user and, at  any moment, allow the user to visualize only the faces still unlabeled.

\textbf{Level of Detail}
As already mentioned, we implement a Level Of Detail (LOD) optimization to enable real-time rendering of large-scale scenarios. For each chunk obtained by splitting the mesh we keep 3 versions at different LOD and dynamically load at high resolution only the meshes within an action range. The user can interact only with the meshes at the highest resolution, those closest to him, thereby significantly alleviating the overall  computational burden. 

\textbf{Labeling Granularity}
A user may require different labeling resolution degrees so as to, e.g., colour either large surfaces or small details.  Therefore, she/he  can choose between a pool of different \emph{weapons} which feature different action ranges, thereby enabling either a more precise or faster labeling. The user chooses the current label  from a color palette (\autoref{fig:teaser}, central picture) and when shooting toward a direction we color each face within the weapon action range of the first hit face. When hitting a face, in Unity we know only the hit face and we must find each face up to a range. As analyzing all faces of the scene can be extremely slow, thus impractical for real-time rendering, we search between the faces belonging only to the same object chunk.

\textbf{Exporting Final Results}
At the end of VR Labeling phase we can save the progress or export the annotated mesh. During export  chunks are merged together into the original mesh. In case the source data was a point cloud, we assign to each 3D point the label of the corresponding voxel.

\subsection{Post-Processing of the Labeled 3D Data}

Once the labeled data have been exported, the tool offers some optional post-processing step. 

\textbf{Multi-Player Integration}
The gamification process potentially enlarge the pool of possible users of our tool. Thus, we can exploit the redundancy of labeling to predict the most confident label for each face or point.
Let us denote  as $\mathcal{S}$ our mesh or point cloud, composed of several elements, $e$, i.e. faces or points respectively. For each point or face $e \in \mathcal{S}$ we define as $\mathcal{H}$  its corresponding histogram of labels, as assigned to it by different players. We can assign to each element $e$ its most confident label by simply finding the most frequent label:

\begin{equation}
	e = argmax(\mathcal{H}), e \in \mathcal{S}
\end{equation}

\textbf{Label Uncertainty}
Since the annotation process by a single user may contain errors, we might wish to assess upon the uncertainty associated with each label. 

Given $n$ annotators we can easily get the label probability distribution $\mathcal{P} = \frac{\mathcal{H}}{n}$ for each element $e$. From the probability distribution we can calculate its entropy:
\begin{equation}
	\mathcal{E}= - \sum p\log{p} 
\end{equation}

The entropy of that distribution can be treated as the uncertainty of the labeling for that element, $u_e$. 
We can leverage this uncertainty to decide which points should be considered noisy ground truth in the annotation process. Moreover, we could exploit it to refine only the high entropy elements both manually or by means of suitable algorithms.

\textbf{Filling}
Some  users may decide to label only partially the whole scenario. Moreover, during the pre-processing we may lose information about few faces where no labels will be available. For these reasons, \algoname{} provides a function for automatic filling missing elements based on their neighborhood. 
We define $\mathcal{S}_{unlabeled}$ the set of elements without any label assigned and $\mathcal{S}_{labeled}$ the set of elements with a label assigned such as $\mathcal{S} = \mathcal{S}_{labeled} \cup{\mathcal{S}_{unlabeled}}$.  Given one point $s_{unlabeled} \in \mathcal{S}_{unlabeled}$, we can find its $K$ closest elements $s_{labeled} \in \mathcal{S}_{labeled}$ and their labels, and build the histogram $H$ of labels of its neighborhood. Then we can easily infer its label:
\begin{equation}
	y=argmax(\mathcal{H})
	\label{eq:argmax}
\end{equation}

In a multi-player setting we can leverage the uncertainty information to further improve the precision and accuracy of the labeling by semantically filling also high uncertainty elements by means of their neighborhood. More precisely, given a fixed uncertainty threshold $th_u$, we consider unlabeled elements that have their uncertainty $u_s$ above $th_u$:

\begin{equation}
	\hat{\mathcal{S}}_{labeled} = \{ s \in \mathcal{S}_{labeled} | u_s > th_u \} 
\end{equation}
\begin{equation}
	\hat{\mathcal{S}}_{unlabeled} = \mathcal{S}_{unlabeled} \cup{}\hat{\mathcal{S}}_{labeled}
\end{equation}

For each $\hat{s}_{unlabeled} \in \hat{\mathcal{S}}_{unlabeled}$ we can find its neighborhood composed of its $K$ closest elements $e_k$. For each $e_k$ we know its labels $y_k$ and its uncertainty information $u_k$. Thus, we can build a weighted histogram of labels for the considered neighborhood, $\mathcal{H}_{u}$, collecting the votes for each label multiplied by their own uncertainty.

At this point the label of $\hat{s}_{unlabeled}$ will be $y=argmax(\mathcal{H}_u)$.

\textbf{Obtaining 2D Segmentations}
We leverage the 3D segmentation to produce 2D segmentations of known RGB images. If a specific set of images comes along its intrinsic and extrinsic camera parameters we can seamlessly render the segmentation through the Blender render engine.

\begin{figure*}[!ht]
    \centering
	\includegraphics[width=\textwidth]{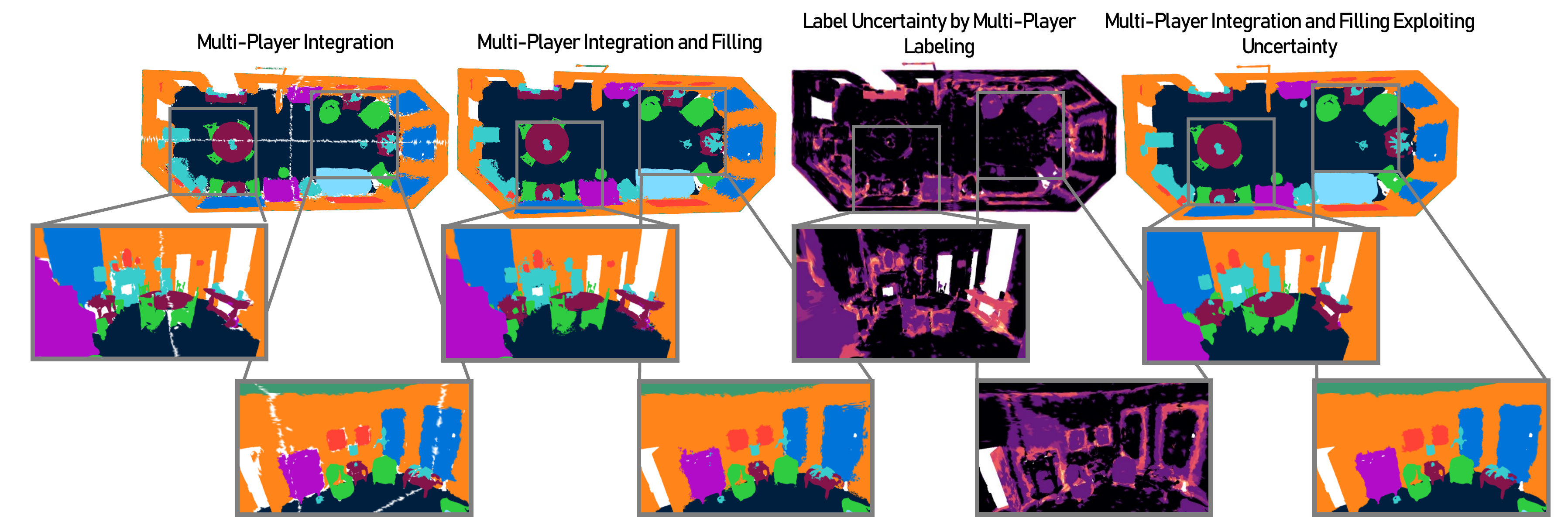}
	\caption{Qualitative comparison of different combination of post-processing steps. From left to right: results after only the multi-player integration step; multi-player integration and filling steps without using label uncertainty; label uncertainty map; multi-player integration and uncertainty aware filling. We can notice that the fourth image has much smoother edges than the second one thanks to  exploitation of the uncertainty map. } 
	\label{fig:confidence}
\end{figure*}

\section{Experimental Results}


\subsection{Efficiency and Accuracy of the Tool}
To evaluate the efficiency and performance of our tool we tested it on the Matterport 3D dataset \cite{chang2017matterport3d}. To perform the evaluation we considered the labeling provided by Matterport as our ground truth. Their labeling has been attained through a series of refinement steps based on several different tools and expertises. They first produced a coarse annotation with a first tool for planar surface labeling, then they used the ScanNet crowd-sourcing interface by Dai et al. \cite{dai2017scannet} to “paint” triangles and name all object instances of the house. Finally a team of 10 expert annotators  refined, fixed and verified the quality of the final labeling. 
In ours tests we labeled an entire Matterport house\footnote{Matterport House ID: \texttt{2t7WUuJeko7}} made out of 6 rooms based on the 13 classes of objects  defined in \cite{couprie2013indoor} (eigen13 categories). We exploit the mapping provided by Matterport from their labeling to eigen13 to obtain ground truths used for testing. They provide face-wise labels since ground truth are meshes. Therefore, we evaluated our results using a mean intersection over union weighted on the area of the faces:

\begin{equation}
    IoU^{c}_{faces} = \frac{A_{TP}}{A_{TP} + A_{FN} + A_{FP}} \quad \text{where} \quad c \in Cl
\end{equation}

\begin{equation}
    mIoU_{faces} = \frac{1}{N}\sum_{c \in Cl} IoU^c_{faces}
\end{equation}

where $Cl$ is the set of classes, $N$ is the total number of classes, ${A_{TP}, A_{FN}, A_{FP}}$ represent the total area of the true positive, false negative and false positive faces respectively for a class ${c}$. 
Furthermore, we provide the percentage of area of faces annotated $A_{labeled}$ over the total area of the labeled ground truth $A_{total}$.

\begin{equation}
    Perc.Area = \frac{A_{labeled}}{A_{total}}\%
\end{equation}

\textbf{Single Player and Multi-Player Integration Results}
We evaluated the annotation of 5 different players without any expertise in 3D modeling. The annotators are 16 to 60 years old, with little or no background in VR systems. We compared their results with the ground truth provided by Matterport. The results are shown in \autoref{tab:singlevsmulti}. The average time needed for the labeling was approximately 2.5 hours. Even though there are users who achieved low labeling performances (player 4 and 5), we notice that integrating results of all players yield the best overall performances of 79.26\% mIoU, surpassing the accuracy of each single user.
As we wanted to analyze what are the most common errors in labeling, we inspected qualitatively each single player results noticing that most frequent errors are correlated with 3D object boundaries and ambiguous object. Therefore, we manually investigate also the GT provided by Matterport finding the same types of errors. In \autoref{fig:gt_errors}, we circled in white the errors on object boundaries while in red the completely mismatching object between our labeling and the Matterport ground truth. In the second row we note that a window (Blue Label) was labeled as a Table (Red Label) in the Matterport GT while with our tool we did not encounter that error. We also found a case of an ambiguous object where a stool has been labeled as an Object (Light Blue Label) by the Matterport ground truth while as a chair by our users (Green Label). These ambiguities and errors in both labeling might be the main cause of achieving lower mIoU score in our labeling and therefore, discarding them with our uncertainty information should lead to better overall performances.

\begin{figure}
	\includegraphics[width=\columnwidth]{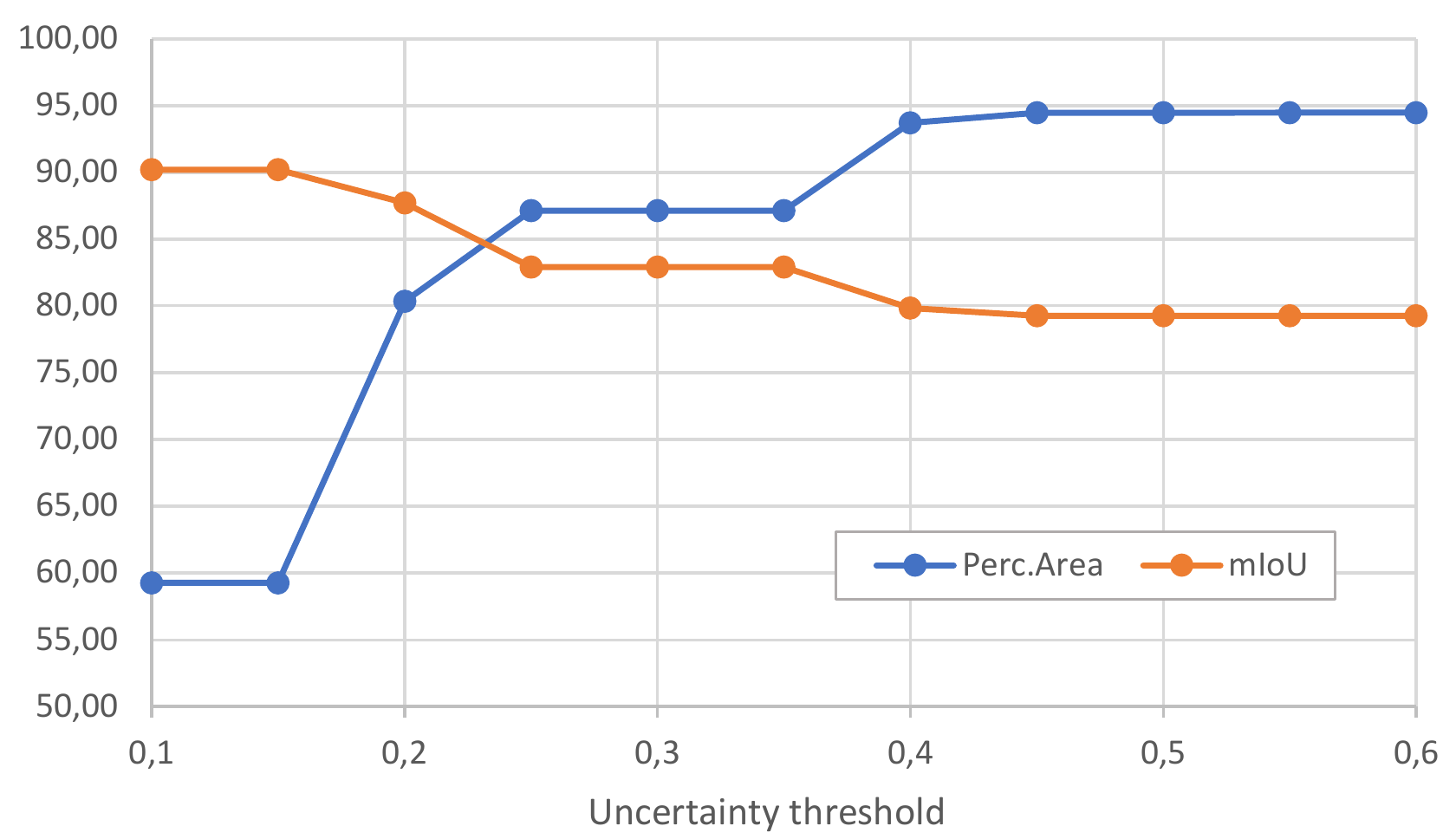}
	\caption{Evaluation at different threshold of uncertainty. The more noisy labels we discard  the higher mIoU wrt Matterport3D ground truth.}
	\label{fig:uncert_graph}
\end{figure}

\textbf{Uncertainty Map Evaluation}
By integrating the results of several users we computed  the label uncertainty  map shown in the third column of \autoref{fig:confidence} and \autoref{fig:gt_errors}. As we wish to evaluate the quality of our labeling,  we analyzed the performances of our labeled mesh at different uncertainty thresholds. In \autoref{fig:uncert_graph} we show the mIoU and the Perc.Area labeled at different uncertainty thresholds. For each threshold we evaluated the mIoU only on the elements with lower uncertainty than the threshold. We see that the higher the threshold the lower the mIoU, symptom of a good uncertainty map. In \autoref{fig:gt_errors} the third column are the uncertainty maps of the labeling where warmer color represents higher uncertainty. We notice that while white circled error are always correlated to high uncertainty, red circled errors might have low uncertainty. This happens because there are object that the majority of the annotators labeled in the same way while they are labeled different in Matterport ground truth  which is correlated to an error in the Matterport ground truth (Windows labeled as a Table).

\textbf{Filling Results w/ or w/o Uncertainty }
\autoref{tab:filling} shows the results of the filling step in various settings. The first five rows report  the results obtained by applying  filling immediately after the single player annotation. These rows  highlight that we were able to fill all the unlabeled faces obtaining Perc.Area of 100\% and  slightly lower  performance. We can see a similar trend also in multi-player integration results where we score  a 76.11\% mIoU while gaining a +5.53\% on the Perc.Area labeled and losing only the 3.15\% in mIoU with respect to the results without  filling.
The last three rows report the results of filling by exploiting the uncertainty map and using different threshold levels, i.e.  0.5, 0.65 and 0.8 respectively, with the best results of 76.13\% mIoU attained with  threshold 0.5. The decreasing trend with higher threshold can be explained thinking that using a higher threshold corresponds with considering good a lot of uncertain elements making filling too difficult and noisy.
\autoref{fig:confidence}  depicts a qualitative comparison between filling strategies. From left to right we can see the labeled mesh before the filling, the mesh filled without uncertainty, the uncertainty map and filling exploiting uncertainty. We highlight how, though the increase in performance in uncertainty aware filling is small, the qualitative results highlight a much smoother labeling.

\begin{figure}
    \centering
	\begin{tabular}{cc}
		\includegraphics[width=0.35\columnwidth]{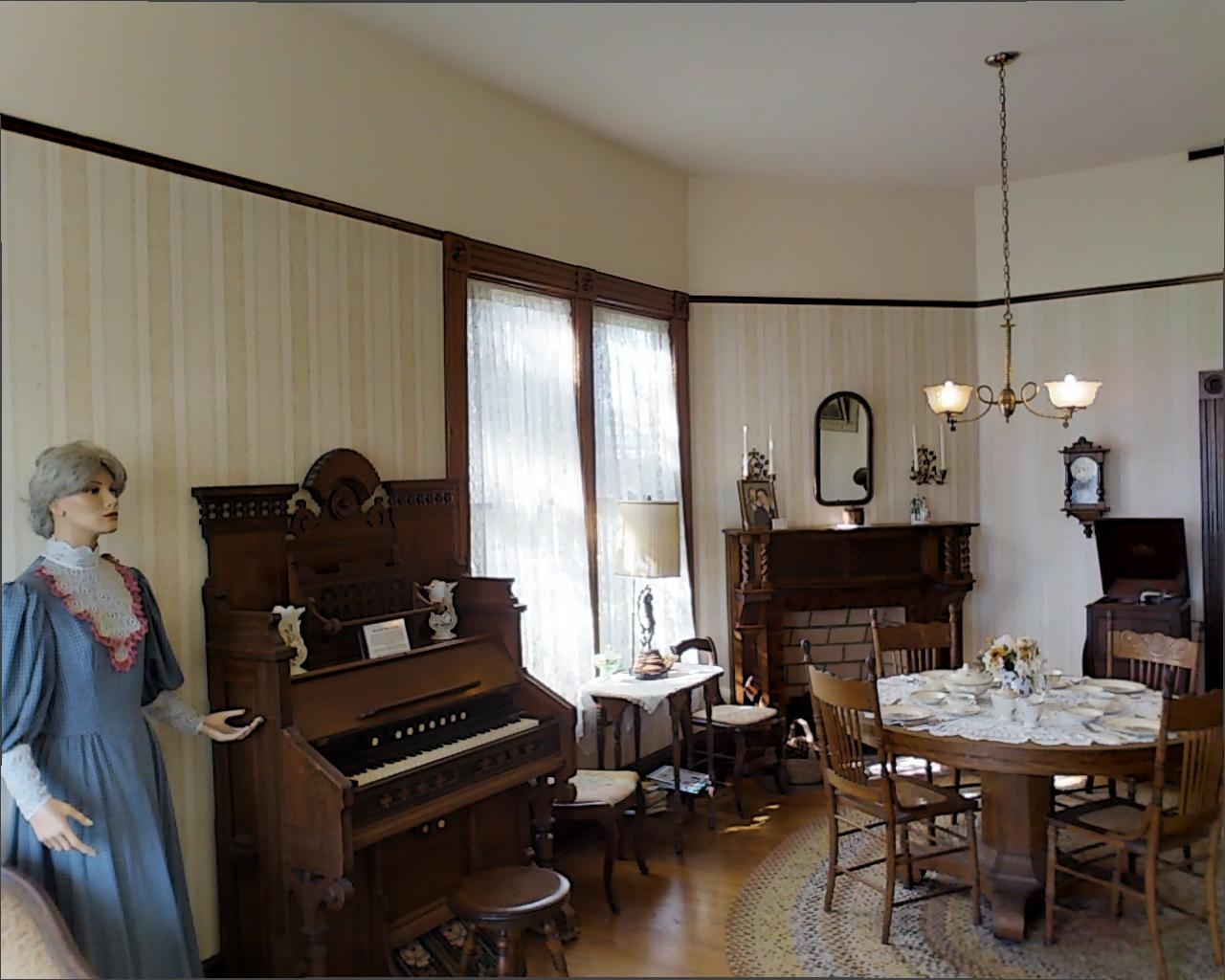} &
		\includegraphics[width=0.35\columnwidth]{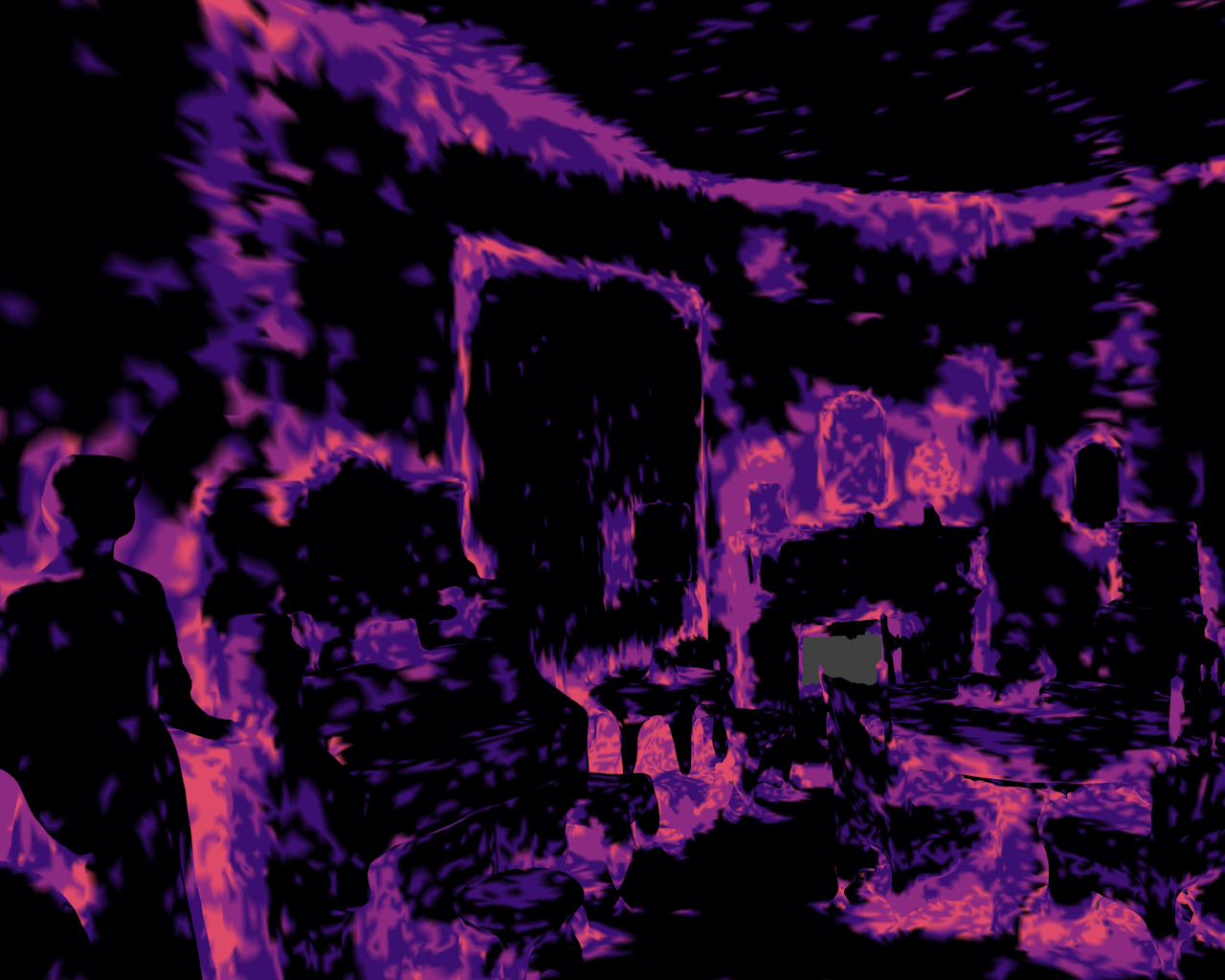} \\
		\includegraphics[width=0.35\columnwidth]{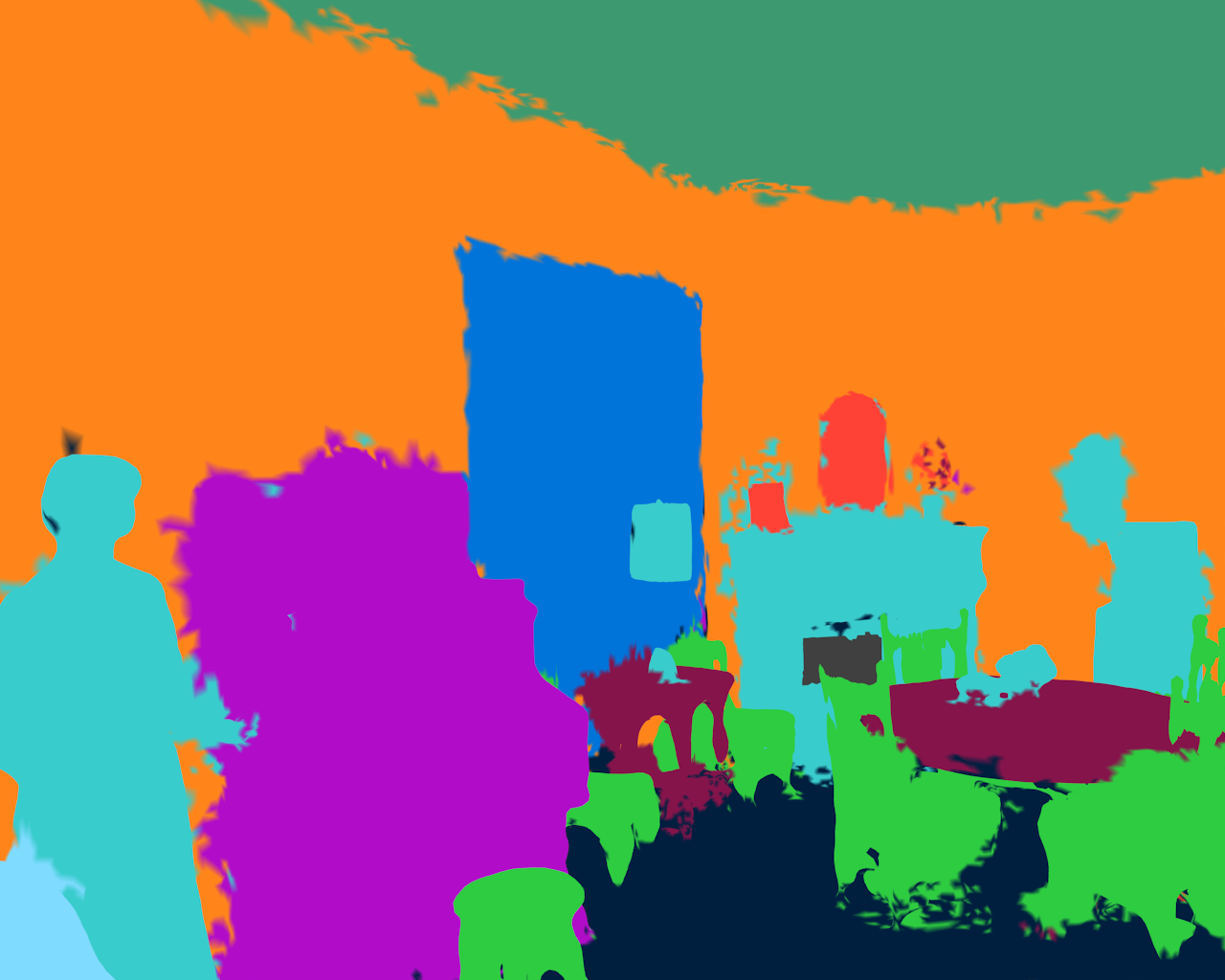} &
		\includegraphics[width=0.35\columnwidth]{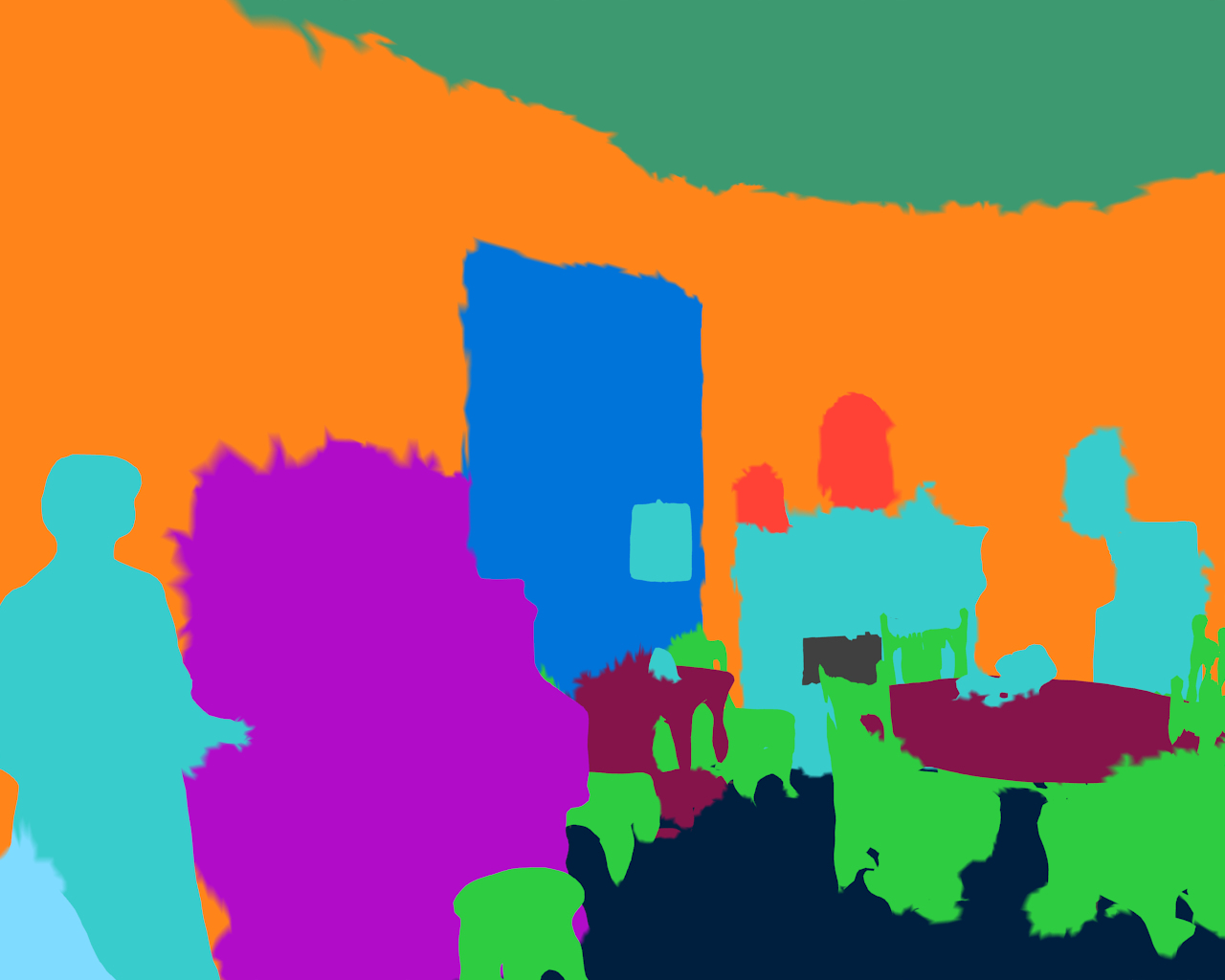}\\
	\end{tabular}
	\caption{Matterport3D: projection of 3D labels into 2D labels. Top Left: RGB image with known camera pose. Bottom Left: 2D labels from a semantically filled mesh without exploiting uncertainty. Top Right: 2D uncertainty map. Bottom Right: 2D labels from a semantically filled mesh by exploiting uncertainty.}
	\label{fig:matterport_qualitatives}
\end{figure}

\textbf{3D to 2D Projection}
\autoref{fig:matterport_qualitatives} illustrates qualitative results of the 2D labels obtained by projecting 3D labels. Given the availability of the RGB image (top left image) and its associated camera intrinsic and extrinsic parameters, we can configure the Blender rendering engine and position the virtual camera in order to obtain the 2D render of the scene. Peculiarly to our tool, we can also provide a 2D uncertainty map by projecting the 3D uncertainty map. Bottom left and right images are rendered from the filled mesh with (right) or without (left) exploiting uncertainty. We can notice that the right image has smoother edges and several  blobs are less noisy. The render took place in approximately 1 second on a GTX 1080 Ti, much less than the hours needed by manual pixel-wise annotation.

\begin{figure*}[t]
	\centering
	\begin{tabular}{ccccc}
		\multicolumn{4}{c}{\includegraphics[width=0.75\textwidth]{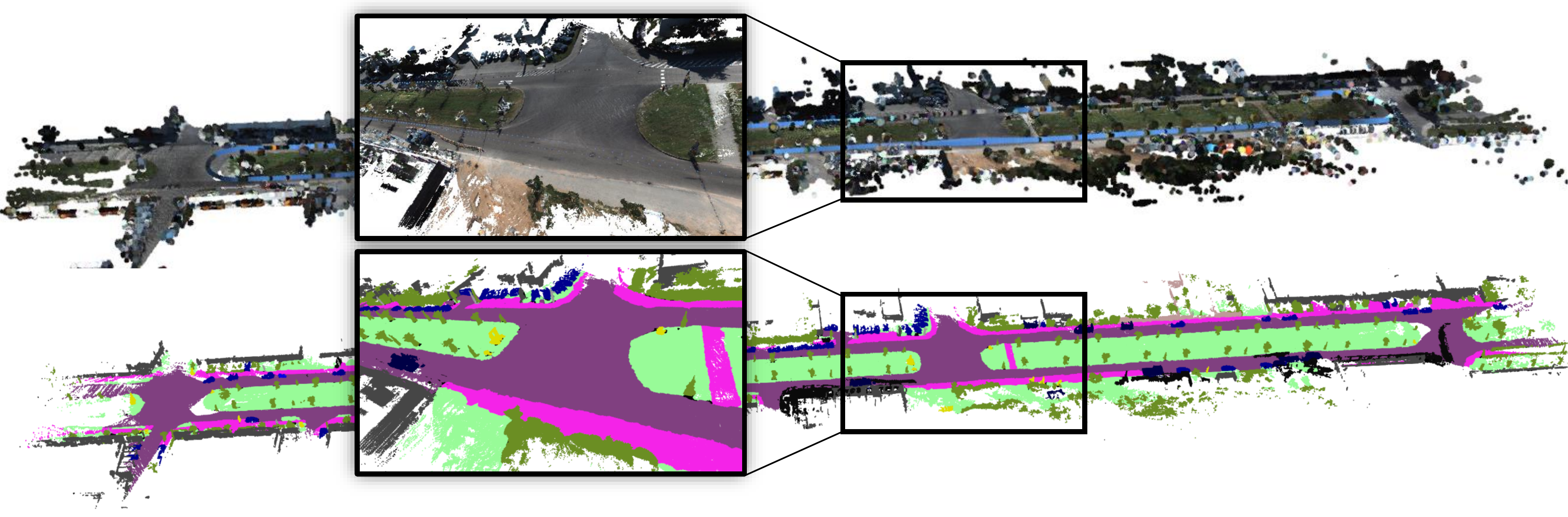}} &
		\includegraphics[width=0.2\textwidth]{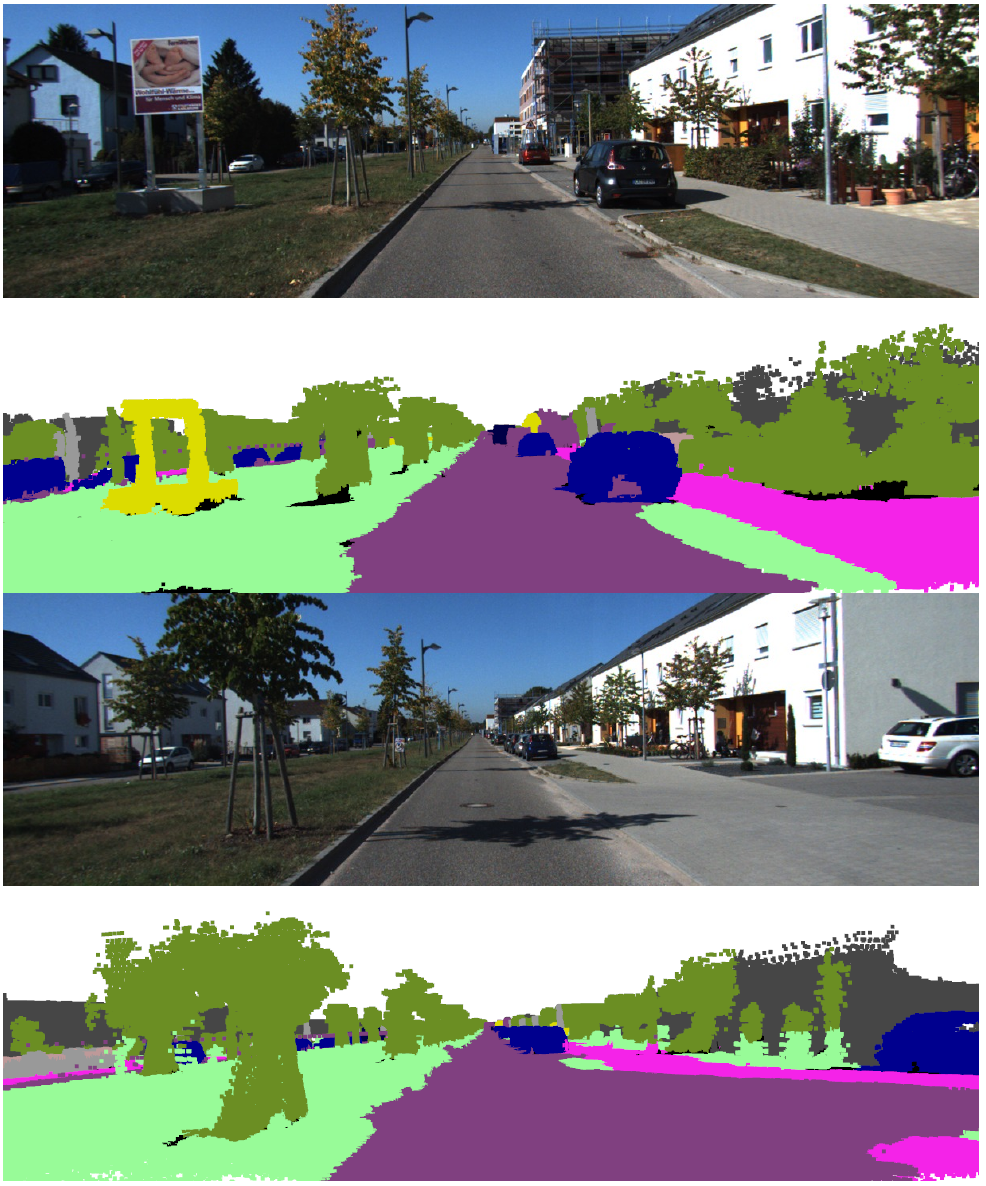}
	\end{tabular}
	\caption{3D and 2D Labeling from a Kitti sequence. Top Left: RGB point cloud. Bottom Left: labeled point cloud obtained by using our tool. Right: RGB images and projected semantic labels}
    \label{fig:pointcloud}
\end{figure*}

\textbf{Large Scale Outdoor Labeling}
We evaluated the effectiveness of our tool in a challenging outdoor scenario: the Kitti Odometry Dataset \cite{geiger2013vision}. We used the provided 3D Lidar data of a static sequence\footnote{Kitti Sequence \texttt{2011\char`_09\char`_30\char`_drive\char`_0020\char`_sync}}, consisting of more than 1000 images equipped with ground truth camera poses. We reconstructed the point cloud, then voxelized  and labeled it by our tool.  Then,  we were able to annotate the whole sequence in approximately 8 hours, a much shorter time compared to other non-VR tool such as \cite{behley2019dataset} which needed about 51 hours for each sequence. Moreover, we could obtain the 2D semantic segmentation associated with the 1000 RGB input images in a few minutes of rendering. \autoref{fig:pointcloud} reports  qualitative results dealing with  our 3D labeling and examples of projected 3D labels.

\section{Conclusions and Future Works}
We have proposed the first 3D semantic labeling tool based on Virtual Reality (VR). Our tool exploits VR alongside with gamification to ease and expedite 3D semantic labeling of large scale scenarios and enlarge the pool of possible annotators to people without any knowledge about 3D modeling. The tool works with the most popular 3D data structures, such as meshes and point clouds.
Moreover, we have shown how to integrate results from multiple users in order to achieve an overall better performance as well as uncertainty map of the labeling process. We have also demonstrated how to integrate the uncertainty map in the labeling process in order to further improve the results. 

We argue that the label uncertainty information may also be leveraged while training deep neural networks for semantic segmentation, e.g. so as to weight the labels in the loss based on their associate uncertainty, as proposed  in some recent works dealing with stereo vision \cite{tonioni2017unsupervised}. Moreover, availability of uncertainty maps may foster the design of novel performance evaluation metrics  which would   take into account the uncertainty of labels.

{\small
\bibliographystyle{IEEEtran}
\bibliography{bibliography}
}

\end{document}